\documentclass[10pt,twocolumn,letterpaper]{article}

\usepackage[pagenumbers]{cvpr} 

\usepackage[accsupp]{axessibility}  

\usepackage{graphicx}
\usepackage{amsmath}
\usepackage{amssymb}
\usepackage{booktabs}
\usepackage{CJKutf8}  
\usepackage{array}
\usepackage{tabularx}
\usepackage{multirow}
\usepackage{subcaption}
\usepackage[dvipsnames]{xcolor}
\usepackage{breakcites} 
\usepackage{comment}
\usepackage{url}
\usepackage{colortbl}
\usepackage{nccmath}
\usepackage{makecell}

\usepackage{times}
\usepackage{epsfig}
\usepackage{mmstyle}

\usepackage{algpseudocode}
\usepackage[ruled]{algorithm2e} 

\SetKwComment{Comment}{/* }{ */}

\DeclareMathAlphabet\mathbfcal{OMS}{cmsy}{b}{n}

\usepackage[colorlinks,pagebackref,breaklinks=true,bookmarks=false]{hyperref}

\usepackage{datenumber}
\usepackage{calc}
\usepackage[mmddyyyy]{datetime}
\newcounter{datetoday}
\newcounter{diffyears}
\newcounter{diffmonths}
\newcounter{diffdays}
\newcommand{\difftoday}[3]{%
      \setmydatenumber{datetoday}{\the\year}{\the\month}{\the\day}%
      \setmydatenumber{diffdays}{#1}{#2}{#3}%
      \addtocounter{diffdays}{-\thedatetoday}%
      \ifnum\value{diffdays}>0
        \def\diffbefore{}%
        \def\diffafter{left}%
      \else
        \def\diffbefore{}%
        \def\diffafter{ago}%
        \setcounter{diffdays}{-\value{diffdays}}%
      \fi
      \setcounter{diffyears}{\value{diffdays}/365}%
      \setcounter{diffdays}{\value{diffdays}-365*\value{diffyears}}%
      \setcounter{diffmonths}{\value{diffdays}/30}%
      \setcounter{diffdays}{\value{diffdays}-30*\value{diffmonths}}%
      \diffbefore
      \ifnum\value{diffyears}=0
      \else
        \ifnum\value{diffyears}>1
            \thediffyears\space years,
        \else
            \thediffyears\space year,
        \fi
      \fi
      \ifnum\value{diffmonths}=0
      \else
        \ifnum\value{diffmonths}>1
            \thediffmonths\space months
        \else
            \thediffmonths\space month
        \fi
      \fi
      \ifnum\value{diffdays}=0
      \else
        \ifnum\value{diffdays}>1
            \thediffdays\space days
        \else
            \thediffdays\space day
        \fi
      \fi
      \diffafter
}

\usepackage[capitalize]{cleveref}
\crefname{section}{Sec.}{Secs.}
\Crefname{section}{Section}{Sections}
\Crefname{table}{Table}{Tables}

\newcolumntype{Y}{>{\centering\arraybackslash}X}

\usepackage{xcolor}
\usepackage{wrapfig}
\usepackage{soul}

\definecolor{yellow}{rgb}{1,1, 0.6}
\definecolor{lightyellow}{rgb}{1,1, 0.8}
\definecolor{orange}{rgb}{1, 0.8, 0.6}
\definecolor{coral}{RGB}{246,131,65}
\definecolor{pinkred}{rgb}{1, 0.6, 0.6}
\definecolor{hotpink}{RGB}{238,64,195}
\definecolor{lavender}{RGB}{207,226,243}
\definecolor{gainsboro}{RGB}{208,224,227}
\definecolor{gainsboro2}{RGB}{217,234,211}
\definecolor{blanchedalmond}{RGB}{252,229,205}

\begin{document}
\begin{CJK}{UTF8}{}
\CJKfamily{mj}

\title{Event Ellipsometer: Event-based Mueller-Matrix Video Imaging}

\author{Ryota Maeda$^{1, 2}$ ~ ~ ~ 
Yunseong Moon$^{1}$ ~ ~ ~
Seung-Hwan Baek$^{1}$ \\[2mm]
$^{1}$POSTECH  ~~~ 
$^{2}$University of Hyogo}

\maketitle



\begin{abstract}
Light-matter interactions modify both the intensity and polarization state of light. Changes in polarization, represented by a Mueller matrix, encode detailed scene information. Existing optical ellipsometers capture Mueller-matrix images; however, they are often limited to capturing static scenes due to long acquisition times. Here, we introduce Event Ellipsometer, a method for acquiring a Mueller-matrix video for dynamic scenes. Our imaging system employs fast-rotating quarter-wave plates (QWPs) in front of a light source and an event camera that asynchronously captures intensity changes induced by the rotating QWPs. We develop an ellipsometric-event image formation model, a calibration method, and an ellipsometric-event reconstruction method. We experimentally demonstrate that Event Ellipsometer enables Mueller-matrix video imaging at 30\,fps, extending ellipsometry to dynamic scenes.
\end{abstract}

\section{Introduction}
\label{sec:intro}
Polarization describes the oscillation of the electric field in light waves, encoding valuable information about the scenes with which light interacts. The polarization state of light can be represented as a Stokes vector $\mathbf{s} \in \mathbb{R}^{4\times1}$~\cite{collett2005field}. Polarimetric image analysis focuses on capturing complete or partial forms of the per-pixel Stokes vector and has been extensively studied for shape-from-polarization~\cite{miyazaki2003polarization,kadambi2015polarized}, diffuse-specular separation~\cite{debevec2000acquiring}, reflection removal~\cite{schechner1999polarization}, seeing through scattering~\cite{schechner2001instant}, and transparent object segmentation~\cite{kalra2020deep}.

Ellipsometry advances polarimetric imaging by analyzing the polarization state of light captured under varying polarization states of illumination. This allows for acquiring polarimetric reflectance, represented as a Mueller matrix $\mathbf{M} \in \mathbb{R}^{4 \times 4}$~\cite{baek2023polarization}, which comprehensively characterizes how light-matter interactions alter the polarization state of incident light. 
Ellipsometry has been widely used in material science~\cite{jellison2018crystallographic, ramsey1994influences}, biology~\cite{arwin2011application, ghosh2011tissue}, and has recently gained attention in computer vision and graphics for 3D shape and reflection analysis~\cite{baek2018simultaneous, baek2022all, scheuble2024polarization}, material acquisition~\cite{baek2020image}, light transport decomposition~\cite{baek2021polarimetric, maeda2024polarimetric}, and photoelasticity analysis~\cite{dave2024nest}.

\begin{figure}[t]
    \centering
    \includegraphics[width=\hsize]{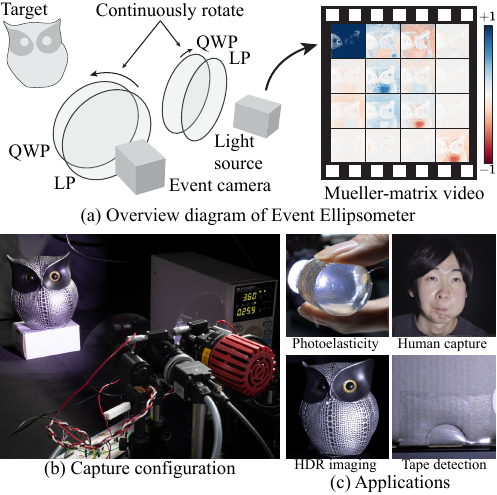}
    \caption{Overview of Event Ellipsometer. (a) Our imaging system captures the Mueller matrix at 30\,fps from event streams induced by the continuously rotating QWPs. (b) With our experimental prototype, (c) we demonstrate ellipsometric analysis for dynamic scenes and various applications.}
    \vspace{-5mm}
    \label{fig:teaser}
\end{figure}

Despite the rich information provided by ellipsometry, its applications have been mainly limited to static scenes because conventional methods require capturing multiple images while mechanically rotating polarizing optics. Typically, more than 20 rotation angles~\cite{baek2020image} are necessary, leading to long acquisition times and making it impractical for capturing dynamic scenes. Although single-shot Mueller matrix imaging techniques using conventional sensors exist~\cite{kudenov2012snapshot, cao2023snapshot, zaidi2024metasurface}, they assume planar target scenes at fixed distances, significantly sacrifice sensor resolution, or use custom-fabricated nano- or micro-optical elements. Additionally, conventional intensity sensors are limited in capturing high-dynamic range (HDR) scenes, often requiring the capture of additional images with multiple exposures.

In this paper, we present \emph{Event Ellipsometer}, a Mueller-matrix imaging method capable of capturing dynamic and HDR scenes. Departing from using conventional intensity sensors, we utilize an event camera, which asynchronously records intensity changes. We equip fast-rotating QWPs and linear polarizers (LPs) in front of both the event camera and an LED light source. The rotating QWPs modulate the polarization state of light emitted and received by the imaging system, resulting in intensity changes captured as events.

We develop an ellipsometric-event image formation model that relates the time differences of adjacent events to the Mueller matrix, normalized by its first element. Using this model, we propose a two-stage Mueller-matrix reconstruction method consisting of per-pixel estimation and spatiotemporal propagation. We incorporate physical validity constraints to handle outliers from sensor noise and scene motion. We also devise calibration methods.

Experimentally, we demonstrate Mueller-matrix imaging at 30\,fps, achieving a mean-squared error of 0.045 for materials with known Mueller matrices. Unlike previous single-shot methods~\cite{zaidi2024metasurface}, Event Ellipsometer can capture non-planar objects and does not compromise spatial resolution. Furthermore, our method is capable of capturing HDR scenes without the need for additional measurement with different exposures.

In summary, our contributions are as follows:
\begin{itemize}
    \item We propose {Event Ellipsometer}, enabling Mueller-matrix imaging for dynamic scenes at 30\,fps.
    \item We develop an imaging system using an event camera and a light source, each equipped with synchronized fast-rotating QWPs.
    \item We formulate an ellipsometric-event image formation model relating event streams to the Mueller matrix, along with a calibration method and a robust ellipsometric-event reconstruction algorithm.
    \item We experimentally show the high accuracy of Event Ellipsometer on samples with known ground truth, and demonstrate ellipsometric analysis of dynamic scenes with applications on photoelasticity, human capture, HDR imaging, and tape detection.
\end{itemize}

\section{Related Work}
\label{sec:related}

\paragraph{Imaging with Polarized Illumination}
Polarimetric imaging with polarized illumination exploits polarization-dependent light transport to extract material and geometric scene properties. Previous methods often capture scenes with single or a few polarized illuminations and fit parametric models to the captured images~\cite{chen2007polarization, treibitz2008active, hwang2022sparse, ichikawa2024spiders}. However, this approach is limited by the representation power of the parametric models, thus cannot reveal the true polarimetric reflectance of real-world scenes.
Ellipsometry extends polarimetric imaging by directly capturing the  Mueller matrix, providing a comprehensive characterization of material polarization properties~\cite{collett2005field, azzam2016stokes, fujiwara2007spectroscopic}. It has applications in material science~\cite{jellison2018crystallographic, ramsey1994influences}, biology~\cite{arwin2011application, ghosh2011tissue}, and has recently been applied to computer vision and graphics for tasks such as 3D shape analysis~\cite{baek2018simultaneous, baek2022all, scheuble2024polarization}, material acquisition~\cite{baek2020image}, and light transport decomposition~\cite{baek2021polarimetric, maeda2024polarimetric}. 
However, conventional ellipsometry techniques are unsuitable for dynamic scenes because they require capturing multiple images while mechanically rotating polarizing optics and capturing each with an intensity sensor, leading to long acquisition. Single-shot Mueller-matrix imaging methods have been proposed~\cite{kudenov2012snapshot, cao2023snapshot, zaidi2024metasurface}, however these approaches often compromise spatial resolution and are limited to planar scenes at fixed distances, restricting their applicability to general dynamic scenes.

\paragraph{Event Cameras for Photometric Analysis}
Event cameras offer a high temporal resolution of microseconds and a HDR compared to standard frame-based intensity cameras~\cite{gallego2020event}. These advantages have been leveraged for photometric analysis by combining continuously modulated optical light sources or filters. For instance, intensity-modulated illumination with an event camera enables radiance estimation~\cite{chen2021indoor, han2023high} and bispectral photometry~\cite{takatani2021event}. Yu et al.~\cite{yu2024eventps} demonstrated event-based photometric stereo for normal estimation with a mechanically-rotating point light source. Hawks et al.~\cite{hawks2022event} and Muglikar et al.~\cite{muglikar2023event} explored passive linear-polarization imaging by rotating a linear polarizer in front of an event camera. Existing event-based vision methods cannot capture full polarization reflectance properties as a Mueller-matrix image, limiting their utility.

\paragraph{Imaging with Rotating Optical Elements}
Rotating optical components have been used for HDR and multispectral imaging~\cite{schechner2004uncontrolled} and high-speed video reconstruction~\cite{chan2023spincam} using a conventional intensity camera. Recently, rotating optics with event cameras have been demonstrated to leverage asynchronous operation and high temporal resolution of the event cameras~\cite{he2024microsaccade, hawks2022event, muglikar2023event}. These methods rotate optical elements only in front of an event camera without illumination modulation. 

\section{Imaging System}
\label{sec:system}

\begin{figure*}[t]
    \centering
    \includegraphics[width=\hsize]{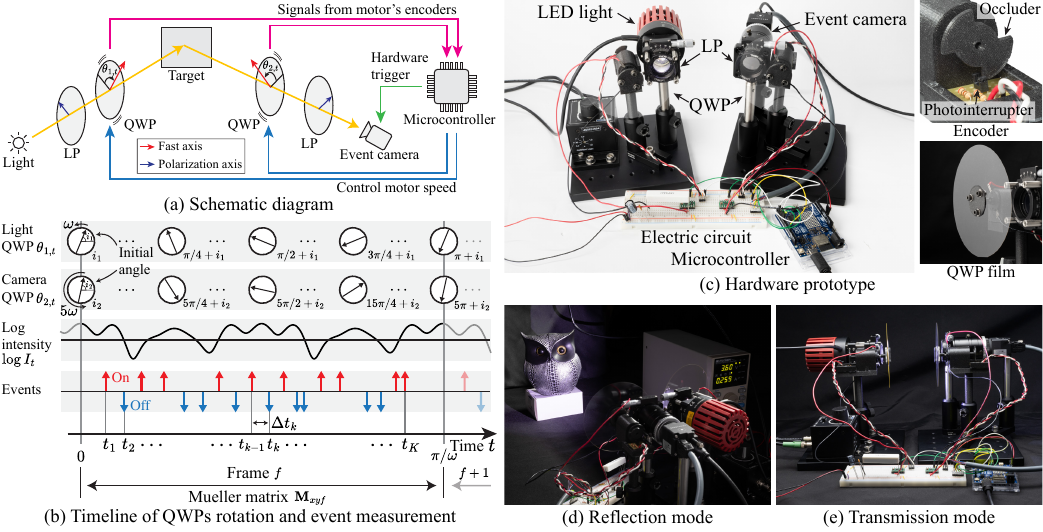}
    \caption{Imaging system of Event Ellipsometer. (a) Schematic diagram illustrating the optical arrangement and hardware operation. (b) Timeline showing the rotation of two QWPs and the event measurement. (c) Our hardware prototype. The system can move the light source and camera position for use in (d) Reflection mode or (e) Transmission mode.}
    \label{fig:prototype}
    \vspace{-5mm}
\end{figure*}

\paragraph{Experimental Prototype}
We build an imaging system using an LED light source (Thorlabs MCWHLP3), fast-rotating QWPs (Edmund Optics WP140HE), fixed LPs (Thorlabs WP25M-VIS), and an event camera (Prophesee EVK4). Figure~\ref{fig:prototype}(a) and (c) depict our setup, where a pair of a QWP and an LP is placed in front of the event camera and the light source, respectively. We rotate the QWPs so that the event camera detects the event streams caused by the rotating QWPs, from which the normalized Mueller matrix is reconstructed. 
We can configure the illumination and the camera modules mounted on different breadboards in both the reflection model and the transmission mode as shown in Figure~\ref{fig:prototype}(d) and (e).
Our system can be seen as a combination of the optical dual rotating retarder~\cite{azzam1978photopolarimetric} and the event-based vision.
For a complete list of components, refer to the Supplemental Material.

\paragraph{Rotating QWPs and Angle Encoder}
We rotate the QWP on the camera side five times faster than the one on the light source side~\cite{smith2002optimization}.
We denote the QWP angles of the light source and the camera as $\theta_{1,t} = \omega t + i_{1}, \theta_{2,t} = 5\omega t + i_{2}$, where $\omega=30 \pi\,\text{rad/sec}$ is the angular velocity of the motor driving the light-source QWP. $i_{1}$ and $i_{2}$ are the initial QWP angles. 
One challenge here is to rotate the QWPs at such high speeds while recording their angles $\theta_{1,t}$ and $\theta_{2,t}$ over time $t$.
To this end, we use two independently controlled brushed direct current (DC) motors rotating the QWP films, resulting in an affordable configuration. Also, we develop custom angle encoders using a 3D-printed occluder, a photointerrupter, and an Arduino microcontroller.
The occluder rotates at the same speed as the QWP and blocks the light path within the photointerrupter at every $\pi$ rotation. The microcontroller detects such change at a microsecond resolution and emits a hardware trigger to the event camera.
Figure~\ref{fig:prototype}(c) shows the angle encoder.

\paragraph{Frame of Mueller Matrix}
Given the rotation speed of $\omega$, every $\pi$ rotation of the light-source QWP takes $\pi/\omega\approx33\,\text{ms}$.
We set this as the effective temporal duration of each frame $f$ of the Mueller-matrix video which we aim to reconstruct.
That is, as shown in Figure~\ref{fig:prototype}(b), each frame $f\in\{1,\cdots,F\}$ of the Mueller matrix is estimated based on the events measured during the temporal duration $\pi/\omega$. $F$ is the number of frames in a reconstructed Mueller-matrix video.

\section{Image Formation}
\label{sec:image_formation}

Here, we relate the event-camera measurements to the scene Mueller matrix which we aim to reconstruct.

\paragraph{Polarimetric Modulation}
The LED light source emits unpolarized light of Stokes vector $\mathbf{s} = [1, 0, 0, 0]^\intercal$. 
At a time $t$, the light passes through the LP and the QWP rotated by angle $\theta_{1,t}$, resulting in the Stokes vector $\mathbf{Q}(\theta_{1,t}) \mathbf{L}(0) \mathbf{s}$, where $\mathbf{Q}(\theta_{1,t})$ and $\mathbf{L}(0)$ denote the Mueller matrices of a QWP rotated by $\theta_{1,t}$ and an LP at $0^\circ$, respectively. The light then interacts with the target scene, undergoing polarization change represented by the scene Mueller matrix $\mathbf{M}$. After interaction, the light passes through another QWP at angle $\theta_{2,t}$ and an LP in front of the event camera, resulting in the time-varying intensity $I_t$ incident on the sensor:
\begin{equation}
    \label{eq:ellipsometer_intensity}
    I_t = [\mathbf{L}(0) \mathbf{Q}(\theta_{2,t}) \mathbf{M} \mathbf{Q}(\theta_{1,t}) \mathbf{L}(0) \mathbf{s}]_0,
\end{equation}
where $[\cdot]_0$ denotes intensity, the first element of the Stokes vector.

We rearrange Equation~\eqref{eq:ellipsometer_intensity} and derive a matrix-vector form:
\begin{equation}
\label{eq:ellipsometer_intensity_matrix}
    I_t = \mathbf{A}_t \hat{\mathbf{M}},
\end{equation}
where $\hat{\mathbf{M}}=[\mathbf{M}_{00}, \mathbf{M}_{01}, \dots, \mathbf{M}_{33}]^\intercal \in \mathbb{R}^{16 \times 1}$ is the vectorized form of $\mathbf{M}$ and $\mathbf{A}_t\in \mathbb{R}^{1 \times 16}$ is the system matrix defined as 
\begin{align}
        \mathbf{A}_t= &[
        1, \alpha_{1}^{2}, \alpha_{1} \alpha_{2}, \alpha_{2}, \alpha_{3}^{2}, \alpha_{1}^{2} \alpha_{3}^{2}, \alpha_{1} \alpha_{2} \alpha_{3}^{2}, \alpha_{2} \alpha_{3}^{2}, \nonumber \\
        &\alpha_{3} \alpha_{4}, \alpha_{1}^{2} \alpha_{3} \alpha_{4}, \alpha_{1} \alpha_{2} \alpha_{3} \alpha_{4}, \alpha_{2} \alpha_{3} \alpha_{4}, -\alpha_{4} \nonumber \\
        &- \alpha_{1}^{2} \alpha_{4}, - \alpha_{1} \alpha_{2} \alpha_{4}, - \alpha_{2} \alpha_{4} ], \text{where} \nonumber  \\
    \alpha_{1} = &\cos{\left(2 i_{1} + 2 \omega t \right)}, 
    \alpha_{2} = \sin{\left(2 i_{1} + 2 \omega t \right)}, \nonumber  \\
    \alpha_{3} = &\cos{\left(2 i_{2} + 10 \omega t \right)},
    \alpha_{4} = \sin{\left(2 i_{2} + 10 \omega t \right)}.
\end{align}
For the full derivation, refer to the Supplemental Document.

\paragraph{Differential Intensity}
The event camera triggers events based on the temporal change of the logarithm of photocurrent~\cite{gallego2020event}. Analytically differentiating the logarithm of Equation~\eqref{eq:ellipsometer_intensity_matrix} with respect to time $t$, we obtain
\begin{align}
    \label{eq:dlogI_ellipsometry}
    \frac{\partial \log I_{t}}{\partial t} 
    = \frac{\frac{\partial I_{t}}{\partial t} }{I_{t}} 
    = \frac{ \frac{\partial \mathbf{A}_t}{\partial t} \hat{\mathbf{M}}}{\mathbf{A}_t \hat{\mathbf{M}}},
\end{align}
where $\frac{\partial \mathbf{A}_t}{\partial t}$ is given as 
\begin{align}
        &\frac{\partial \mathbf{A}_t}{\partial t}=[
                0, - 4 \alpha_{1} \alpha_{2}, 2 \alpha_{1}^{2} - 2 \alpha_{2}^{2}, 2 \alpha_{1}, - 20 \alpha_{3} \alpha_{4}, \nonumber \\
                & -20 \alpha_{1}^{2} \alpha_{3} \alpha_{4} - 4 \alpha_{1} \alpha_{2} \alpha_{3}^{2}, 2 \alpha_{1}^{2} \alpha_{3}^{2} - 20 \alpha_{1} \alpha_{2} \alpha_{3} \alpha_{4} - 2 \alpha_{2}^{2} \alpha_{3}^{2}, \nonumber \\
        &2 \alpha_{1} \alpha_{3}^{2} - 20 \alpha_{2} \alpha_{3} \alpha_{4}, 10 \alpha_{3}^{2} - 10 \alpha_{4}^{2},  \nonumber \\
        & 10 \alpha_{1}^{2} \alpha_{3}^{2} - 10 \alpha_{1}^{2} \alpha_{4}^{2} - 4 \alpha_{1} \alpha_{2} \alpha_{3} \alpha_{4}, \nonumber \\
        &2 \alpha_{1}^{2} \alpha_{3} \alpha_{4} + 10 \alpha_{1} \alpha_{2} \alpha_{3}^{2} - 10 \alpha_{1} \alpha_{2} \alpha_{4}^{2} - 2 \alpha_{2}^{2} \alpha_{3} \alpha_{4}, \nonumber \\
        & 2 \alpha_{1} \alpha_{3} \alpha_{4} + 10 \alpha_{2} \alpha_{3}^{2} - 10 \alpha_{2} \alpha_{4}^{2}, - 10 \alpha_{3}, - 10 \alpha_{1}^{2} \alpha_{3} + 4 \alpha_{1} \alpha_{2} \alpha_{4}, \nonumber \\
        & - 2 \alpha_{1}^{2} \alpha_{4} - 10 \alpha_{1} \alpha_{2} \alpha_{3} + 2 \alpha_{2}^{2} \alpha_{4}, - 2 \alpha_{1} \alpha_{4} - 10 \alpha_{2} \alpha_{3}].
\end{align}

\paragraph{Events from Polarimetric Modulation}
For a pixel, we denote the time difference between consecutive events as $\Delta t_k$, where $k \in \{1,\cdots,K\}$ is the event index in each frame $f$. $K$ is the number of detected events at that pixel in the frame $f$. 
The time difference is known to be related to the change of photocurrent according to Taylor's expansion~\cite{gallego2020event}:
\begin{equation}
    \label{eq:dlogI_event}
    \frac{\partial \log I_{t_k}}{\partial t} = \frac{p_k C}{\Delta t_k},
\end{equation}
where event polarity $p_k \in \{+1, -1\}$ is the sign of intensity change, and $C$ is a constant threshold of the event camera.

By combining Equations~\eqref{eq:dlogI_ellipsometry} and \eqref{eq:dlogI_event}, we relate the measured per-pixel events to the Mueller matrix, resulting in the image formation model:
\begin{align}
    \label{eq:image_formation}
    &\frac{p_k C}{\Delta t_k} = \frac{ \frac{\partial \mathbf{A}_{t_k}}{\partial t_k} \hat{\mathbf{M}}}{\mathbf{A}_{t_k} \hat{\mathbf{M}}} \nonumber \\
    & \Rightarrow { \left( \frac{\partial \mathbf{A}_{t_k}}{\partial t_k} - \frac{p_k C}{\Delta t_k} \mathbf{A}_{t_k}\right) } \hat{\mathbf{M}} = 0 \nonumber \\
    & \Rightarrow \mathbf{B}_{t_k} \hat{\mathbf{M}} = 0,
\end{align}
where $\mathbf{B}_{t_k} \in \mathbb{R}^{1\times16}$ is the system matrix.

\section{Reconstruction}
\label{sec:reconstruction}
\begin{figure}[t]
    \centering
    \includegraphics[width=\linewidth]{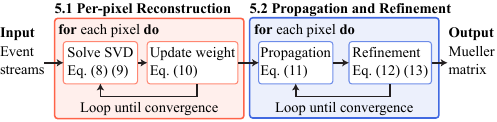}
    \caption{Overview of our Mueller-matrix reconstruction pipeline. This method consists of two steps: (1) per-pixel reconstruction and (2) propagation and refinement.}
    \label{fig:algorithm_overview}
    \vspace{-3mm}
\end{figure}

The image formation model in Equation~\eqref{eq:image_formation} allows us to reconstruct the Mueller matrix $\hat{\mathbf{M}}$. Our reconstruction method, depicted in Figure~\ref{fig:algorithm_overview}, consists of two main steps: (1) per-pixel reconstruction and (2) spatio-temporal propagation.

\begin{figure*}[t]
    \centering
    \includegraphics[width=\hsize]{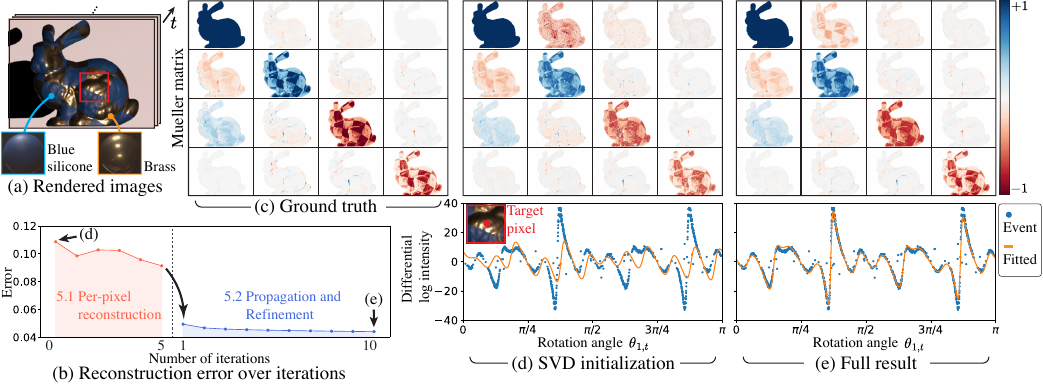}
    \caption{Synthetic data evaluation result. (a) The rendered images include two materials: blue silicone and brass. (b) The plot shows the error (mean absolute error) of the reconstructed Mueller matrix over the number of iterations. (c) Ground truth Mueller matrix. (d)\&(e) Top: the reconstructed Mueller-matrix images for the SVD initialization and the full-stage, respectively. {Pixels with insufficient event counts are visualized in white.} Bottom: a plot of the differentiation of log intensity and fitted line with our method.}
    \label{fig:synthetic_result}
    \vspace{-3mm}
\end{figure*}

\subsection{Per-pixel Reconstruction}
\label{subsec:per-pixel_reconstruction}

In this step, our goal is to estimate a Mueller matrix in a computationally efficient manner and being robust to noisy events. 

First, we stack the vectors $\mathbf{B}_{t_k}$ for all $k$, forming a matrix $\mathbf{B} \in \mathbb{R}^{K \times 16}$. We then solve for the vectorized Mueller matrix $\hat{\mathbf{M}}$ by minimizing the weighted least-squares problem:
\begin{equation}
\label{eq:least_square}
    \underset{\hat{\mathbf{M}}}{\text{minimize}} \| \mathbf{W} \mathbf{D} \mathbf{B} \hat{\mathbf{M}} \|_2^2,
\end{equation}
where $\mathbf{W} = \operatorname{diag}(\mathbf{w})$ is a diagonal matrix of weights $\mathbf{w} = [w_1, \dots, w_K] \in \mathbb{R}^{1\times K}$, initialized as ones: $w_k = 1$. The matrix $\mathbf{D} = \operatorname{diag}([\Delta t_1,\dots,\Delta t_K])$ is a weight matrix that adjusts the influence of each event according to sampling density. This weighting ensures that more densely sampled events at small $\Delta t$ do not dominate the model fitting.
We solve Equation~\eqref{eq:least_square} using singular value decomposition (SVD) to obtain $\hat{\mathbf{M}}$ while avoiding the trivial zero solution.

Second, we filter the reconstructed Mueller matrix $\hat{\mathbf{M}}$ to ensure physical validity. Sensor noise and rapid scene dynamics can make solving Equation~\eqref{eq:least_square} challenging, often resulting in physically invalid Mueller matrices. These invalid matrices can trap the optimization in local minima. To address this, we apply Cloude's Mueller matrix filtering~\cite{cloude1990conditions} to project the estimated matrix onto the space of physically valid Mueller matrices:
\begin{equation}
    \label{eq:filter}
    \hat{\mathbf{M}} \leftarrow \text{CloudeFilter}(\hat{\mathbf{M}}),
\end{equation}
where $\text{CloudeFilter}(\cdot)$ denotes the filtering function.

Third, we recalculate the weight vector $\mathbf{w}$ using the physically valid Mueller matrix $\hat{\mathbf{M}}$:
\begin{equation}
    w_k = {1}/{\max ( | \mathbf{B}_{t_k} \hat{\mathbf{M}} |, \epsilon )},
\end{equation}
where $\epsilon$ is a small positive constant to avoid division by zero sets as $10^{-6}$ in our experiments. This weight reduces the influence of outlier events and enables robust estimation.

We iterate these three steps—solving the weighted least-squares problem, applying physical validity filtering, and recomputing the weights—until convergence, resulting in the Mueller matrix estimate $\hat{\mathbf{M}}$ for each pixel $(x,y)$ and frame $f$. We set the maximum number of iterations to 5 in our experiments

\subsection{Spatio-temporal Propagation and Refinement}
\label{subsec:propagation_and_refinement}
Here, we start by denoting the estimated Mueller matrix at a pixel $(x,y)$ and frame $f$ as $\hat{\mathbf{M}}_{xyf}$ and the corresponding system matrix as $\mathbf{B}_{xyf}$, and weight matrix as $\mathbf{D}_{xyz}$.
This second stage refines the Mueller matrix $\hat{\mathbf{M}}_{xyf}$ estimated from the first stage using spatio-temporal neighboring pixels. This process is inspired by PatchMatchStereo from stereoscopic depth estimation~\cite{barnes2009patchmatch, bleyer2011patchmatch, galliani2015massively}.

First, we perform propagation. We consider neighboring pixels in both spatial and temporal axes, denoted by $(x', y', f') \in \mathcal{N}(x, y, f)$. The set $\mathcal{N}$ defines a spatio-temporal neighborhood extended from the red-black spatial-only neighborhood~\cite{galliani2015massively}. For each pixel at a frame $(x, y, f)$, we update its Mueller matrix if a neighboring pixel $(x', y', f')$ offers a lower error:
\begin{align}
&\hat{\mathbf{M}}_{xyf} \gets \hat{\mathbf{M}}_{x'y'f'}, \nonumber \\
    &\text{if} \quad \| \mathbf{D}_{xyz} \mathbf{B}_{xyf} \hat{\mathbf{M}}_{xyf} \|_1 >  \| \mathbf{D}_{xyz} \mathbf{B}_{xyf} \hat{\mathbf{M}}_{x'y'f'} \|_1.
\end{align}

Next, we refine the Mueller matrix by applying random perturbations and accepting the perturbation if it reduces the error:
\begin{align}
&\hat{\mathbf{M}}_{xyf} \gets \hat{\mathbf{M}}_{xyf}^{\text{perturb}}, \text{where} \nonumber \\
    &\hat{\mathbf{M}}_{xyf}^{\text{perturb}} = \text{CloudeFilter}(\hat{\mathbf{M}}_{xyf} \odot \left( \mathbf{1} + \sigma \mathbf{N} \right)),\\
    &\text{if} \quad \| \mathbf{D}_{xyz} \mathbf{B}_{xyf} \hat{\mathbf{M}}_{xyf} \|_1 >  \| \mathbf{D}_{xyz} \mathbf{B}_{xyf} \mathbf{M}_{xyf}^{\text{perturb}} \|_1, 
\end{align}
where $\hat{\mathbf{M}}_{xyf}^{\text{perturb}}$ is the perturbed Mueller matrix, $\odot$ denotes element-wise multiplication, $\mathbf{1}$ is an all-ones matrix, $\mathbf{N} \in \mathbb{R}^{4 \times 4}$ contains random values from a standard normal distribution, and $\sigma$ is a scalar set to $0.01$. This multiplicative perturbation preserves the relative scale of the Mueller matrix elements, allowing larger elements to undergo proportionally larger adjustments while smaller elements change minimally. We apply Cloude's filtering to the perturbed Mueller matrix to ensure physical validity.

We repeat the propagation and refinement steps iteratively for a fixed number of iterations set to 10 in our experiments to obtain the final Mueller matrix $\hat{\mathbf{M}}_{xyf}$.

\begin{figure*}[t]
    \centering
    \includegraphics[width=\hsize]{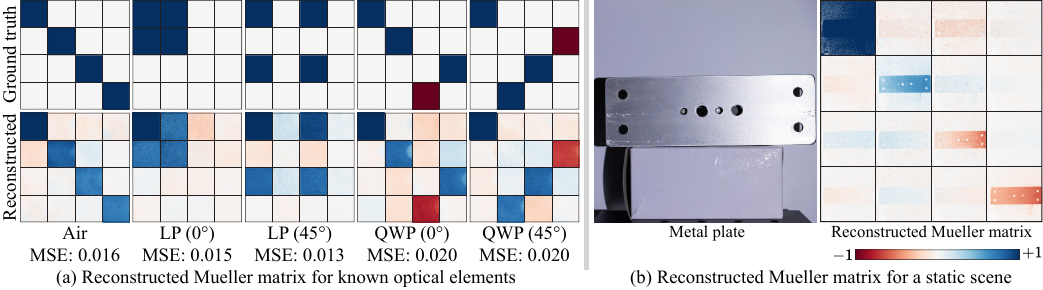}
    \caption{Assessment of reconstructed Mueller matrix on real data. (a) Evaluation with known optical elements. {We show the corresponding mean squared errors (MSEs).} (b) Measurement on an in-the-wild metal plate induces strong diagonal components.}
    \label{fig:result_assessment}
\end{figure*}

\section{Calibration}
We perform one-time calibration of the system parameters: contrast threshold ($C$) and QWP offset angles ($i_1$ and $i_2$). Calibration details are provided in the Supplementary Material.

\paragraph{Contrast Threshold}
We calibrate the per-pixel contrast threshold $C$ of an event sensor by analyzing pixel responses under controlled illumination stimuli. We increase and decrease the LED intensity linearly over time, and the event camera directly captures the LED light, generating events. We then fit the contrast threshold $C$ to the events per each pixel using Equation~\eqref{eq:dlogI_event}. 

\paragraph{QWP Offset Angle}
We obtain the offset angles of the fast axis of the QWP in our system: $i_1$ and $i_2$ by capturing another QWP sample with known fast axis. Given the known Mueller matrix of the reference QWP, we find the best offset angles that minimize the reconstruction error of Equation~\eqref{eq:dlogI_event}. 

\section{Results}
\label{sec:results}

\paragraph{Implementation}
We implemented our reconstruction algorithm in C++ using OpenMP CPU parallelization.  
It takes 62 seconds for per-pixel reconstruction (Section~\ref{subsec:per-pixel_reconstruction}), and 68 seconds for the propagation and refinement processes (Section~\ref{subsec:propagation_and_refinement}), measured for processing 500$\times$500-resolution event image sequence of 30 frames, with an average event rate of 153 MEv/s, on the AMD Ryzen 7 7800X3D 8-core processor. 


\paragraph{Validation on Synthetic Data}
To evaluate our method, we render synthetic data in three steps.
First, we mimic our capture system by replacing the event camera with an intensity camera in Mitsuba3~\cite{Mitsuba3}, and render an image sequence of a synthetic scene for each QWP angle with interval $\omega t=0.01^\circ$.
We use the real-world polarimetric BRDFs to construct the synthetic scene~\cite{baek2020image}. 
Then, the rendered images are converted to event streams by simulating event-camera processing based on DVS-Voltmeter~\cite{lin2022dvs}. 
Last, we add Gaussian noise with a standard deviation of 0.5 and replace 5\% of true events with values sampled from a Gaussian distribution with a standard deviation of 5.0. Figure~\ref{fig:synthetic_result} shows the synthetic data and reconstructed Mueller matrix using our method. While initial SVD-based estimation suffers from noise with a reconstruction error of 0.11, our full reconstruction improves accuracy with a reconstruction error of 0.04.

\paragraph{Validation on Real Data}
Figure~\ref{fig:result_assessment}(a) shows that our reconstruction closely matches the corresponding pseudo ground truth of real-world samples: air, linear polarizer ($0^\circ$ and $45^\circ$), quarter-wave-plate ($0^\circ$ and $45^\circ$). Figure~\ref{fig:result_assessment}(b) shows the result on an in-the-wild sample: a metal plate. The strong diagonal components in this result indicate the preservation of polarization, which is characteristic of reflections from metal surfaces. 
Notably, these Mueller matrix images are captured in just 33ms, significantly faster than the conventional ellipsometer, which requires several minutes due to its frame-based imaging principle.

\begin{figure*}[t]
    \centering
    \includegraphics[width=1.0\hsize]{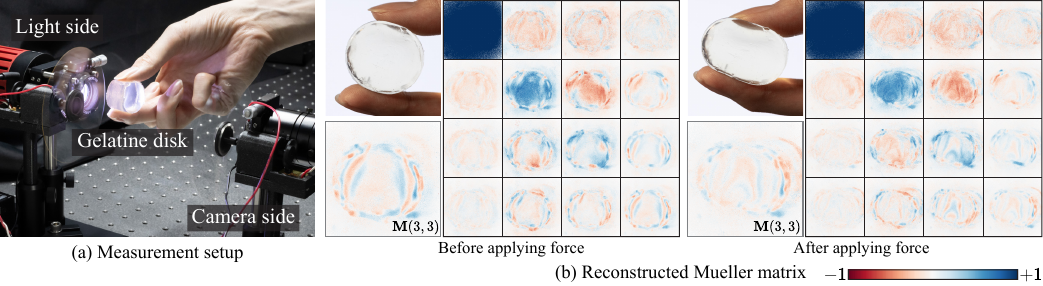}
    \caption{Photoelasticity analysis. (a) Experimental setup for measuring a gelatine disk in transmission mode. We gradually apply force for the gelatine disk to observe changes in photoelastic properties. (b) Reconstructed Mueller matrix images with different forces, revealing complex stress-dependent polarimetric patterns. 
    {Notably, as force increases, the Mueller matrix image shows denser fringe patterns.}}
    \label{fig:result_photoelasticity}
    \vspace{-3mm}
\end{figure*}

\paragraph{Photoelasticity of Transparent Gelatine}
Photoelasticity is an optical property whereby dielectric materials exhibit birefringence under deformation. Imaging photoelasticity has applications in mechanical stress and material analysis~\cite{scafidi2015review,wang2017mechanical}. 
Analyzing Mueller-matrix images reveals such stress distributions even in transparent objects~\cite{dave2024nest}. Our system enables observing photoelasticity at video rates. Figure~\ref{fig:result_photoelasticity} shows Mueller-matrix images of a gelatine disk, revealing stress-dependent fringe patterns.

\begin{figure}[t]
    \centering
    \includegraphics[width=\hsize]{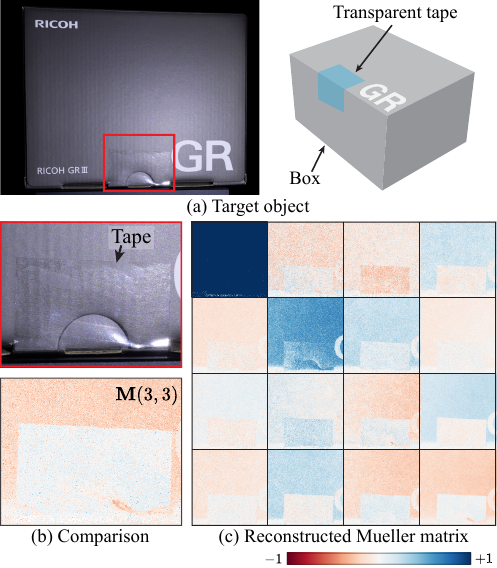}
    \caption{Transparent tape detection. (a) The target object is transparent sticky tape on a box. (b)\&(c) The Reconstructed Mueller matrix reveals the tape region, which is difficult to see with a conventional RGB camera.}
    \vspace{-3mm}
    \label{fig:result_tape}
\end{figure}

\paragraph{Transparent Tape Detection}
Detecting transparent objects is challenging for conventional cameras. Sticky tape, a common transparent material used for sealing cardboard, exhibits birefringent properties due to the molecular structure of its stretched plastic. Figure~\ref{fig:result_tape} shows the measurement result of sticky tape on a cardboard box. The reconstructed Mueller matrix shows the birefringent property, and we can clearly recognize the tape region. This demonstration shows the potential for inspecting dynamically moving sealed boxes in automated industrial processes.

\begin{figure*}[t]
    \centering
    \includegraphics[width=\hsize]{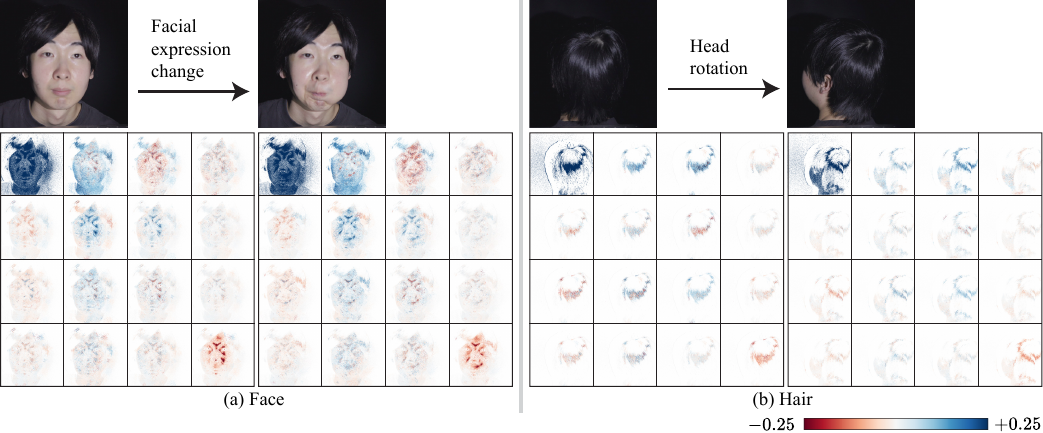}
    \caption{Mueller matrix acquisition for capturing dynamic human (a) face and (b) hair, demonstrating the capture of diffuse and specular polarimetric responses.}
    \label{fig:result_human}
    \vspace{-3mm}
\end{figure*}

\begin{figure}[t]
    \centering
    \includegraphics[width=\hsize]{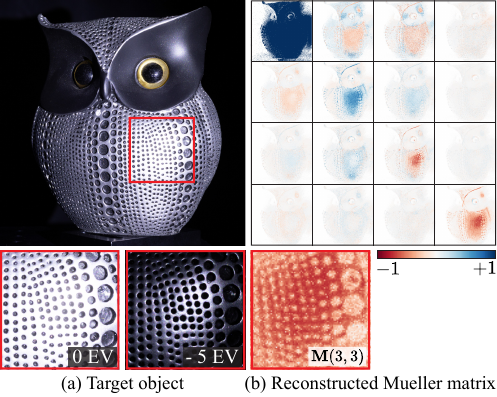}
    \caption{Mueller matrix measurement for a HDR scene. (a) The target scene contains regions with both strong specular reflection and black diffuse reflection.  (b) Our proposed method achieves an accurate Mueller-matrix image for the HDR scene in just 33 ms.}
    \label{fig:result_hdr}
    \vspace{-3mm}
\end{figure}

\paragraph{Dynamic Human Capture}
Ellipsometers, as a comprehensive polarization imaging technique for human capture, can reveal hidden properties of the human face and hair that are difficult to analyze in conventional imaging and non-ellipsometric polarization imaging~\cite{debevec2000acquiring, azinovic2023high}.
In Figure~\ref{fig:result_human}, we demonstrate the Mueller-matrix reconstruction of dynamic facial expressions and hair movements, revealing intricate polarization properties.
The reconstructed Mueller matrix for the face exhibits strong diagonal components associated with specular reflection, while the {non-diagonal} regions display weakly polarized reflection with a dependency on surface normal {along the face edges}. The hair result shows polarization property in the specular highlight regions.

\paragraph{HDR Mueller-matrix Imaging}
Our event-based ellipsometer enables Mueller-matrix imaging of HDR scenes. Figure~\ref{fig:result_hdr} shows reconstruction results on a scene with specular and dark diffuse reflections. While conventional ellipsometers require multiple exposures to prevent overexposure, our method captures the Mueller matrix in HDR scenes at 30\,fps without the need for additional measurement.

\section{Discussion}
\label{sec:discussion}

\paragraph{Normalized Mueller-Matrix Imaging}
Normalized Mueller matrix has been extensively used as it provides rich scene information~\cite{baek2018simultaneous,baek2021polarimetric,dave2024nest}. 
Since event cameras only detect intensity changes, not intensity itself, our method also reconstructs a normalized version of Mueller matrix per pixel, $\mathbf{M}/\mathbf{M}(0,0)$. 
We leave reconstructing full Mueller-matrix as a future work that might be accomplished using an event-intensity hybrid imaging system. 

\paragraph{End-to-end Real-time Pipeline}
Although our capture speed is fast, our current Mueller-matrix reconstruction relies solely on CPU acceleration, resulting in a non-real-time pipeline from capture to reconstruction. Implementing GPU acceleration could enable on-the-fly capture, reconstruction, and visualization~\cite{yu2024eventps,galliani2015massively}.

\paragraph{Event Bandwidth and Sensitivity}
The ideal setting of an event camera is to have a low contrast threshold and a short refractory period, to capture subtle intensity changes. However, this produces many events, exceeding the transmission bandwidth, resulting in lost or delayed events~\cite{gallego2020event}. 
In contrast, when the contrast threshold is high, weakly polarized phenomena cannot be detected as events.
Given this trade-off, we defined a region of interest {(ROI)} in the event sensor and empirically configured the camera settings to find the best parameters. 

\paragraph{DC Motor Synchronization}
For faster rotation speed and precise synchronization between two motors, it is a viable alternative to use brushless DC motors, also known as an electronically commutated (EC) motors. However, this increases system building costs.

\paragraph{Gradient {Descent}}
Gradient descent in an automatic differentiation framework offers an alternative approach for Mueller-matrix reconstruction.
However, the number of events differs across pixels, which complicates the construction of a dense tensor for efficient computation on modern GPUs. 

\section{Conclusion}
\label{sec:conclusion}
We have introduced Event Ellipsometer, a Mueller-matrix video imaging method that combines an event camera and a light source with fast-rotating QWPs. Our image formation model, calibration, and reconstruction method enable robust Mueller-matrix imaging at 30\,fps. We validate our method on synthetic and real data, and demonstrate Mueller-matrix imaging on photoelasticity, dynamic human hair and face capture, HDR imaging, and transparent tape detection.

\paragraph{Acknowledgments}
Seung-Hwan Baek was supported by Korea NRF (RS-2023-00211658, RS-2024-00438532, 2022R1A6A1A03052954, RS-2024-00437866), and Samsung Research Funding \& Incubation Center of Samsung Electronics under Project Number SRFC-IT1801-52.

\clearpage
{\small
\bibliographystyle{ieeenat_fullname}
\bibliography{references}
}

\end{CJK}
\end{document}


\begin{CJK}{UTF8}{}
\CJKfamily{mj}

\title{Event Ellipsometer: Event-based Mueller-Matrix Video Imaging \\
- Supplemental Document -\\
}

\author{Ryota Maeda$^{1, 2}$ ~ ~ ~ 
Yunseong Moon$^{1}$ ~ ~ ~
Seung-Hwan Baek$^{1}$ \\[2mm]
$^{1}$POSTECH  ~~~ 
$^{2}$University of Hyogo}

\maketitle

In this supplemental document, we provide additional results and details of Event Ellipsometer. 

\tableofcontents

\newpage

\section{Image Formation}

\subsection{Mueller Matrices of Optical Elements}
Mueller matrix of linear polarizer $\textbf{L}$ and quarter-wave plate $\textbf{Q}$ with rotation angle $\theta$ are represented by as follows~\cite{collett2005field}:
\begin{equation}
    \mathbf{L}(\theta) = 
    \frac{1}{2}
    \left[\begin{matrix}
        1 & \cos{\left(2 \theta \right)} & \sin{\left(2 \theta \right)} & 0 \\
        \cos{\left(2 \theta \right)} & \cos^{2}{\left(2 \theta \right)} & \sin{\left(2 \theta \right)} \cos{\left(2 \theta \right)} & 0 \\
        \sin{\left(2 \theta \right)} & \sin{\left(2 \theta \right)} \cos{\left(2 \theta \right)} &  \sin^{2}{\left(2 \theta \right)} & 0 \\
        0 & 0 & 0 & 0
    \end{matrix}\right],
\end{equation}
\begin{equation}
    \mathbf{Q}(\theta) = 
    \left[\begin{matrix}
        1 & 0 & 0 & 0 \\
        0 & \cos^{2}{\left(2 \theta \right)} & \sin{\left(2 \theta \right)} \cos{\left(2 \theta \right)} & - \sin{\left(2 \theta \right)} \\
        0 & \sin{\left(2 \theta \right)} \cos{\left(2 \theta \right)} & \sin^{2}{\left(2 \theta \right)} & \cos{\left(2 \theta \right)} \\
        0 & \sin{\left(2 \theta \right)} & - \cos{\left(2 \theta \right)} & 0
    \end{matrix}\right].
\end{equation}

\subsection{Time Difference Computation with Refractory Time}
The event camera has a refractory time $\eta$, during which the sensor is unable to detect intensity changes. To incorporate this time into the reconstruction process, we compute the time differences $\Delta t_k$ as follows:
\begin{equation}
    \Delta t_k = t_{k+1} - t_k - \eta. 
\end{equation}
The refractory time $\eta$ reduces the time difference and the impact of the refractory time $\eta$ is negligible when the naive time difference ($t_{k+1}{-}t_k$) is large but becomes significant for the smaller time difference. We obtained the refractory time value using Metavision SDK~\footnote{Metavision SDK, https://docs.prophesee.ai/stable/index.html}. 

\subsection{QWP Angles}
As introduced in the main paper, we denoted the QWP angles as $\theta_{1,t} = \omega t + i_1$ and $\theta_{2,t} = 5\omega t + i_2$, and the detailed definition of $i_1$ and $i_2$ are defined as follows:
\begin{align}
    \label{eq:i_1}
    i_1 &= \phi_\mathrm{calib1}, \\
    \label{eq:i_2}
    i_2 &= \phi_\mathrm{calib2} - 5 \phi_{\mathrm{dynamic}, f},
\end{align}
where $\phi_\mathrm{calib1}$ and $\phi_\mathrm{calib2}$ are fixed offset angles determined in one-time calibration (Section~\ref{sec:calib_qwp_offset}). $\phi_{\mathrm{dynamic}, f}$ denotes the dynamic offset angle between the two QWPs, measured for each frame (Section~\ref{sec:angle_time_correspondence}). 

\section{Reconstruction Method}

\begin{figure}[h]
    \centering
    \includegraphics[width=\hsize]{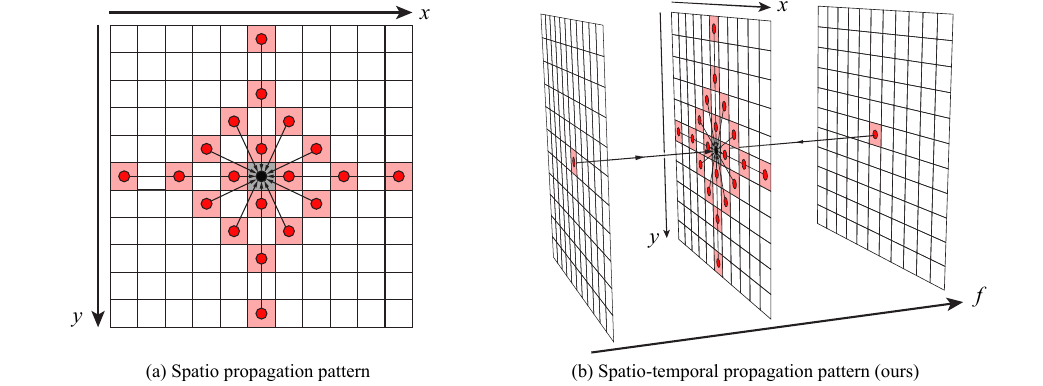}
    \caption{Spatial-temporal propagation pattern. The center pixel (depicted in black) is updated using Mueller-matrix candidates from neighboring pixels (depicted in red). (a) The original red-black propagation pattern, which considers only the spatial axes ($x$ and $y$)~\cite{galliani2015massively}. (b) Our propagation pattern extends this scheme into the temporal axis ($f$).}
    \label{fig:propagation_pattern}
\end{figure}

\subsection{Spatio-temporal Propagation Pattern}
Our spatial-temporal propagation scheme is inspired by the red-black pattern for stereo depth estimation~\cite{galliani2015massively}, which enables efficient parallel computation. Extending this spatial pattern into the temporal domain, we define a spatio-temporal pattern as illustrated in Figure~\ref{fig:propagation_pattern}. This pattern preserves simultaneous updates for both red and black pixels, ensuring computational efficiency across spatial and temporal dimensions in video processing.

\section{Hardware Implementation}
This section describes the details of our hardware implementation.

\subsection{Part List}
Table~\ref{tab:parts_list} shows the parts list of our hardware prototype.

\begin{table}[h]
    \centering
    \begin{tabular}{lll} \toprule
        \textbf{Part description} & \textbf{Model name} & \textbf{Quantity} \\ \midrule
        Quarter-wave plate & Edmund Optics WP140HE & 2 \\
        Linear polarizer & Thorlabs WP25M-VIS & 2 \\
        Event camera & Prophesee EVK4 & 1 \\
        Lens & 8mm F1.4 & 1\\
        LED light & Thorlabs MCWHLP3 & 1 \\ 
        LED driver & Thorlabs LEDD1B & 1 \\
        Microcontroller & Arduino UNO R4 & 1 \\
        Motor & Maxon A-max 19 (249993) & 2 \\ 
        Motor driver & Pololu DRV8835 & 2 \\  
        Photointerrupter & SHARP GP1S094HCZ0F & 2 \\
        DC power supply & BK Precision 9111 & 1 \\
        \bottomrule
    \end{tabular}
    \caption{Parts list of our hardware prototype.}
    \label{tab:parts_list}
\end{table}


\subsection{Rotation Speed Control}
\label{sec:rotation_speed_control}
To ensure constant-speed rotation of the QWPs, we implemented a feedback-based control mechanism using the measured rotation speed measured with photointerrupters. Each photointerrupter consists of an LED and a photodetector, which work together with an occluder attached to the motor shaft. This setup detects interruptions in the light path, which are then converted into rotation speed measurements. Based on these measurements, we regulate the motor current through pulse-width modulation (PWM), with the duty cycle controlled by a Proportional-Integral-Derivative (PID) controller. We manually tuned the PID controller parameters to achieve stable rotation.

\subsection{Rotation Angle and Event Time Correspondence}
\label{sec:angle_time_correspondence}
\begin{figure}[h]
    \centering
    \includegraphics[width=\linewidth]{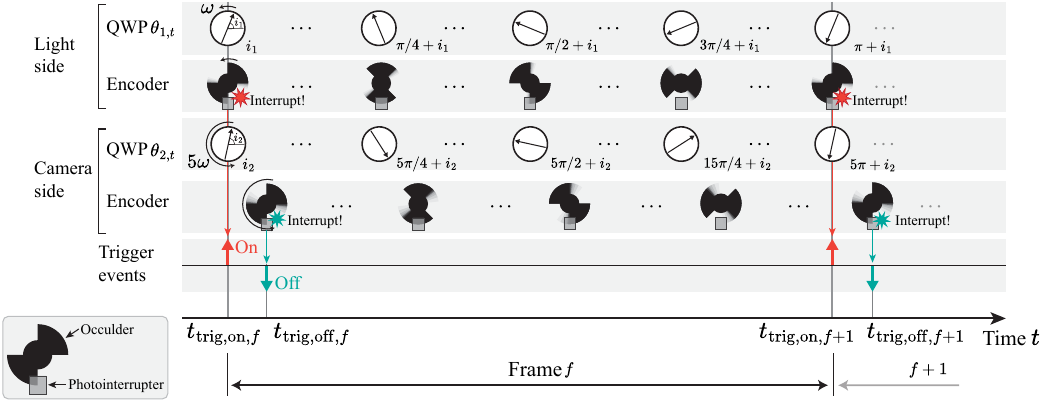}
    \caption{Timeline of motor rotation and trigger events.}
    \label{fig:timeline_trigger}
\end{figure}

To establish the correspondence between the rotation angle and the recorded event time, we utilized the trigger interface feature provided in Prophesee EVKs. This interface allows for the recording of trigger events generated by external hardware signals. Utilizing this functionality, we recorded the interrupted timing from the motor’s encoder and incorporated it into the Mueller-matrix reconstruction process. As shown in Figure~\ref{fig:timeline_trigger}, we recorded two types of trigger event times for each frame $f$: $t_{\mathrm{trig, on}, f}$ corresponding to the light-side encoder, and $t_{\mathrm{trig, off}, f}$ corresponding to the camera-side encoder. These trigger times are utilized in two ways: converting event time to correspondence rotation angle and calculating the dynamic offset angles between two QWPs. To convert event times to corresponding angles, we used the following relation: 
\begin{equation}
    \theta = \omega t = \pi \frac{t - t_{\mathrm{trig, on}, f}}{t_{\mathrm{trig, on}, f+1} - t_{\mathrm{trig, on}, f}}.
\end{equation}
The dynamic offset angles between the two QWPs are calculated as follows:
\begin{equation}
    \phi_{\mathrm{dynamic}, f} = \pi \frac{t_{\mathrm{trig, on}, f} - t_{\mathrm{trig, off}, f}}{t_{\mathrm{trig, on}, f+1} - t_{\mathrm{trig, on}, f}}.
\end{equation}
These calculations ensure the alignment of rotation angles with event times on independently rotated motors.

\subsection{Contrast Threshold Calibration}
\begin{figure}[h]
    \centering
    \includegraphics[width=\linewidth]{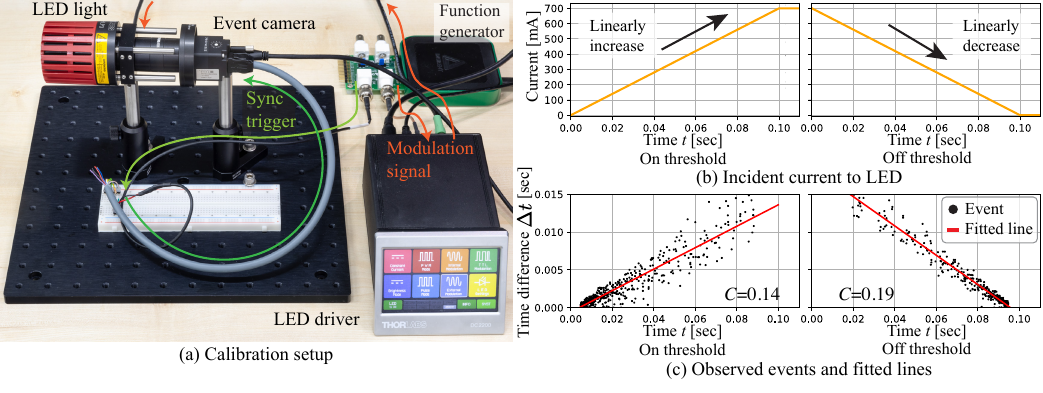}
    \caption{Calibration of contrast threshold. (a) Calibration setup. (b) We linearly increase/decrease the incident current to the LED, and the light intensity is proportionally modulated by the amount of current. (c) From observed events, we fit the linear equation and get the per-pixel contrast threshold for both on and off events. This figure shows one example of a fitted result resulting in $C{=}0.14$ for on events and $C{=}0.19$ for off events.}
    \label{fig:calib_thresh}
\end{figure}

We calibrate the contrast threshold $C$ using the events observed under controlled illumination stimuli. As shown in Figure~\ref{fig:calib_thresh}(a),  the event camera directly faced the LED light. The light intensity was modulated using an LED driver (Thorlabs DC2200) and a function generator (Digilent Analog Discovery 3), which also provided a synchronization trigger to align event timings with modulation.
We employ the linearly increasing/decreasing function to modulate the intensity as follows:
\begin{equation}
    I_t = a t + b,
\end{equation}
where $a$ is coefficient of slope, and $b$ is initial intensity of light. Figure~\ref{fig:calib_thresh}(b) shows the electronic current to LED in our calibration. 
The derivative of the logarithmic intensity under this stimulus is given by:
\begin{equation}
    \frac{\partial \log I_t}{\partial t} = \frac{\frac{\partial I_t}{\partial t}}{I_t} = \frac{a}{at+b}.
\end{equation}
This establishes a linear relationship between $t$ and $\Delta t$ as follows
\begin{align}
    & \frac{pC}{\Delta t} = \frac{a}{at+b}  \\
    & \Rightarrow {\Delta t = pC \left(t + \frac{b}{a}\right)}.
\end{align}
Using this equation, we can calibrate the contrast threshold $C$ by linear regression with respect to $t$ and $\Delta t$.

Calibration was performed for both "on" and "off" events for each pixel. To reduce noise, measurements were repeated 15 times, and all events were used. Additionally, to avoid saturation due to an excessive number of events, we restricted the region of interest (ROI) to $80{\times}80$ pixels, with the ROI scanned across the entire pixel array. Figure~\ref{fig:calib_thresh}(c) shows the fitted result on one pixel.

\subsection{QWP Offset Angle Calibration}
\label{sec:calib_qwp_offset}
\begin{figure}[h]
    \centering
    \includegraphics[width=\hsize]{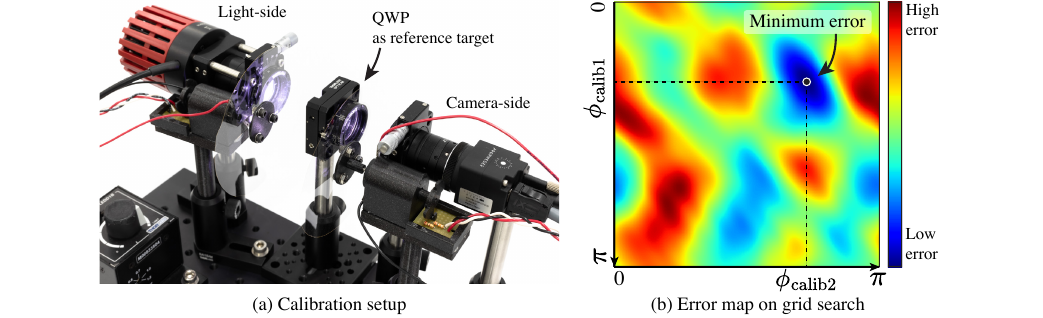}
    \caption{Calibration of QWP offset angles. (a) Calibration setup. We placed the QWP between light and camera in transmission mode. (b) The error map to find the optimal angles with the grid search algorithm. This error map shows the unique solution with the measurement of QWP.}
    \label{fig:calib_qwp_offset}
\end{figure}

We calibrate the offset angles of the fast axis of the QWP, denoted as $\phi_\mathrm{calib1}$ and $\phi_\mathrm{calib2}$ (see Equation~\eqref{eq:i_1} and \eqref{eq:i_2}). These offsets arise during the attachment of the QWP and the occluder to the motor's shaft. To calibrate these angles, we measure the events of a QWP as a known reference target, as shown in Figrue~\ref{fig:calib_qwp_offset}(a). Unlike air or a linear polarizer, the use of a QWP enables the determination of a unique solution. The modulated intensity under this condition is expressed as follows:
\begin{equation}
    \label{eq:ellipsometer_intensity_qwp}
    I_t = [\mathbf{L}(0) \mathbf{Q}(\theta_{2,t}) \mathbf{D}(\alpha) \mathbf{Q}(\pi/4) \mathbf{Q}(\theta_{1,t}) \mathbf{L}(0) \mathbf{s}]_0,
\end{equation}
where $\mathbf{D}$ is the Mueller matrix of depolarizer, representing depolarization due to non-ideal optics:
\begin{equation}
    \mathbf{D}(\alpha) = 
    \left[\begin{matrix}
        1 & 0 & 0 & 0 \\
        0 & \alpha & 0 & 0 \\
        0 & 0 & \alpha & 0 \\
        0 & 0 & 0 & \alpha
    \end{matrix}\right].
\end{equation}
The parameter $\alpha$ ($0 \leq |\alpha| \leq 1$) is the depolarization factor, with smaller values of $\alpha$ corresponding to greater depolarization. In this calibration, $\alpha$ was set to 0.8.

To find the offset angles, we used the log-derivative form of Equation~\eqref{eq:ellipsometer_intensity_qwp} and performed a grid search to identify the pair of angles that best fit the observed events. Figrue~\ref{fig:calib_qwp_offset}(b) shows the error map that highlights unique angle combinations.

\section{Additional Results}

\subsection{Ablation Study}
In addition to Figure~4(c) in the main paper, we conducted an additional ablation study on real data of air (Figure~5(a)).
First, we evaluate the impact of Cloude’s filter in the per-pixel reconstruction step. With and without Cloude’s filter, the reconstruction error is 0.03697 and 1.12×$10^{12}$, respectively. This result indicates the strong effectiveness of the filtering step in improving reconstruction accuracy.
Second, we assess the impact of propagation and random perturbation, as summarized in Table~\ref{tab:ablation}. Combining both methods improves reconstruction performance. Although the effect of random perturbation is relatively subtle, it plays a crucial role in preventing direct value copying from neighboring pixels, thereby enabling more precise optimization.

\begin{table}[h]
    \centering
    \caption{Ablation study of propagation and random perturbation with mean squared error (MSE) and mean absolute error (MAE).}
    \begin{tabular}{cccc}
        \toprule
        \textbf{Propagation} & \textbf{Perturbation} & \textbf{MSE} ($\downarrow$) & \textbf{MAE} ($\downarrow$) \\
        \midrule
        -- & \checkmark & 0.03614 & 0.12781 \\
        \checkmark & -- & 0.01665 & 0.08816 \\
        \checkmark & \checkmark & \textbf{0.01661} & \textbf{0.08688} \\
        \bottomrule
    \end{tabular}
    \label{tab:ablation}
\end{table}

\subsection{Rotation Stability of the Prototype}
\begin{figure}[h]
    \centering
    \includegraphics[width=\hsize]{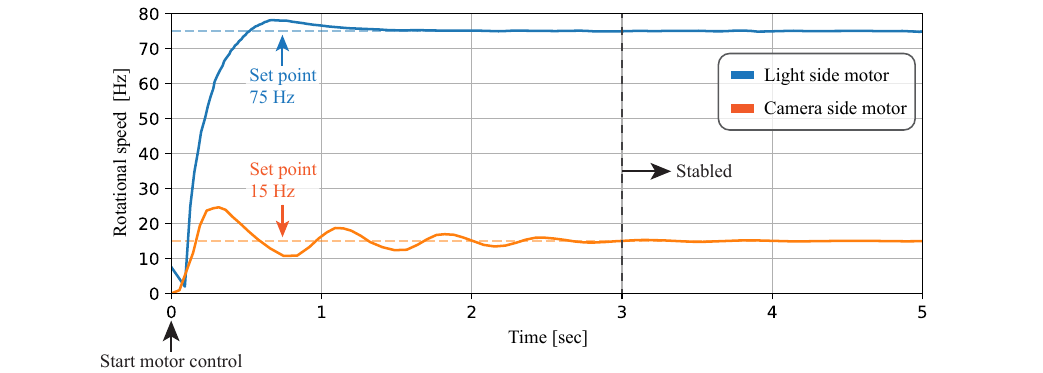}
     \caption{Measured rotation speed of the two motors. The set point is 15 Hz {with a standard deviation of 0.31} for the light-side motor and 75 Hz {with a standard deviation of 0.07} for the camera-side motor. The rotation speeds stabilize at the set point within three seconds.}
    \label{fig:result_rotation_speed}
\end{figure}

To assess the rotation stability of the motors in our prototype, we recorded the encoder interruption times and subsequently converted them into rotation speeds. Figure~\ref{fig:result_rotation_speed} shows the measured rotation speeds of the two motors, set to 15Hz and 75Hz, respectively. Both motors achieve stable rotation at their target speeds within three seconds.

\begin{figure}[h]
    \centering
    \includegraphics[width=\hsize]{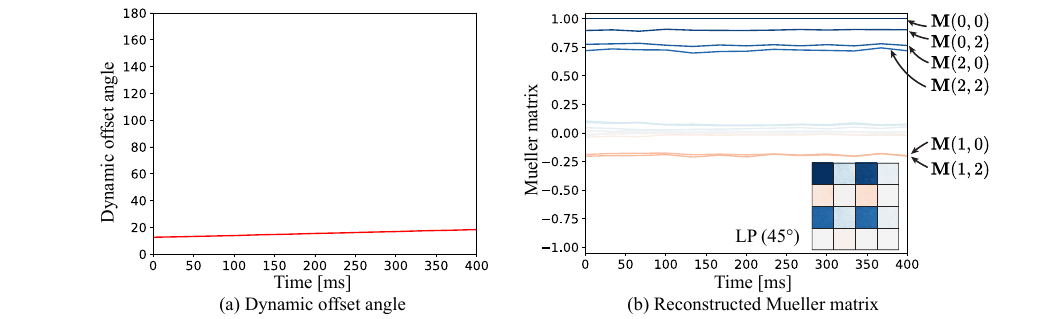}
    \caption{Synchronization and reconstruction stability.}
    \label{fig:rotation_stability}
\end{figure}

Additionally, we examined the temporal stability of Mueller matrix reconstruction. Figure~\ref{fig:rotation_stability} shows the dynamic offset angle $\phi_{\mathrm{dynamic}, f}$ and the reconstructed Mueller matrix of LP$45^{\circ}$ (shown in Figure 5(a)). While slight variations exist due to imperfect synchronization, the reconstructed Mueller matrix remains stable. 

\subsection{Comparison with Frame-based Method}

\begin{figure}[h]
    \centering
    \includegraphics[width=\hsize]{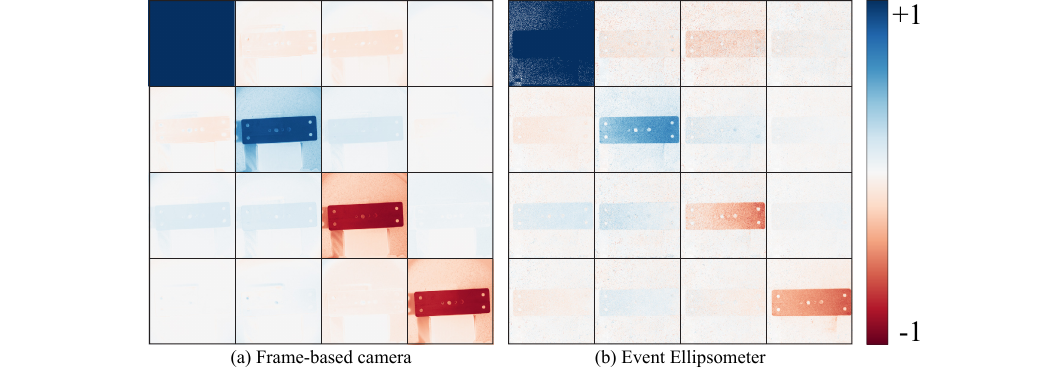}
    \caption{Evaluation with the frame-based method~\cite{baek2020image} requires 5~min to capture, whereas our approach achieves it in just 33 ms.}
    \label{fig:metal_plate_rebuttal}
\end{figure}

We conducted an additional evaluation by comparing our method with a conventional frame-based approach ~\cite{baek2020image} for the metal plate scene (Figure~4(b)). For the frame-based measurement, we captured exposure-bracketed images by changing the exposure times, which were merged into a HDR image. HDR images were captured under 36 combinations of QWP rotation angles, requiring approximately 5 minutes to acquire a single Mueller matrix image. As shown in Figure~\ref{fig:metal_plate_rebuttal}, our event-based method produces results comparable to the frame-based method, validating its accuracy while achieving significantly faster capture times (33~ms). Inevitable differences arise due to various variations, such as camera positioning.

\subsection{Comparison with Single-shot Method}
\begin{figure}[h]
    \centering
    \includegraphics[width=\hsize]{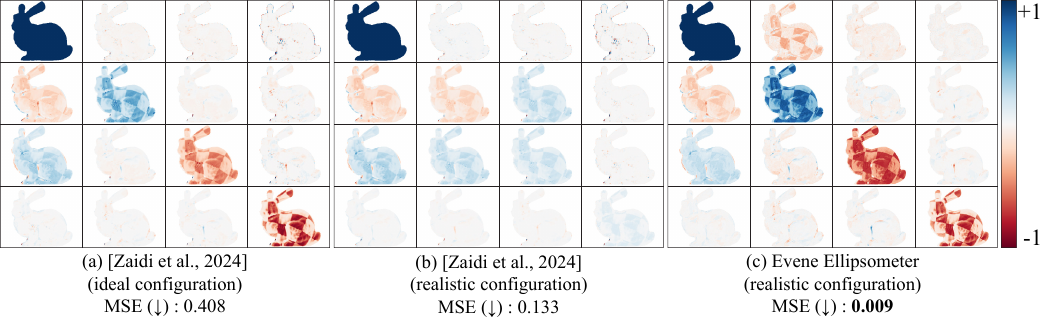}
    \caption{Comparison with the single-shot method~\cite{zaidi2024metasurface} using synthetic data.}
    \label{fig:single_shot}
\end{figure}

As shown in Figure~\ref{fig:single_shot}, we compare our method against a single-shot method~\cite{zaidi2024metasurface} using synthetic data. This single-shot method relies on specialized and high-cost nano-photonic metasurfaces, which simultaneously alter both ray positions and directions, as well as polarization states, making the simulation with arbitrary 3D geometry challenging.

To enable a feasible simulation, we simplify the target scene's geometry to a planar surface. With the ideal coaxial setup of the camera and light source, we can ensure exact correspondence between camera pixels and the projection pattern. The reconstructed Mueller matrix image under this ideal setup is shown in Figure~\ref{fig:single_shot}(a).
However, their method exhibits high sensitivity to misalignment under realistic conditions (Figure~\ref{fig:single_shot}(b)). In contrast, our method ( Figure~\ref{fig:single_shot}(c)) can reconstruct the scene while allowing greater flexibility in light positioning, enhancing practical usability. 

\subsection{Statistics of Event Rate}

We computed event rate statistics across all scenes. The minimum, maximum, average, and standard deviation of event rates is 10.3, 171.9, 91.8, and 41.1~MEv/s, respectively. Empirically, we confirmed that stable event capture within this event rate range is achieved by appropriately setting the ROI.

\subsection{Detailed Results with Decomposed Mueller Matrix}

Figure~\ref{fig:result_photoelasticity_supp} to \ref{fig:result_tape_supp} show the more detailed result of the reconstructed Mueller matrix in the main paper along with decomposed images that offer deeper insights into the scene. These decompositions are obtained through the Lu-Chipman decomposition method~\cite{lu1996interpretation}, which decomposes the reconstructed Mueller matrix $\mathbf{M}$ into three distinct Mueller matrices:
\begin{equation}
    \mathbf{M} = \mathbf{M}_{\Delta} \mathbf{M}_{R} \mathbf{M}_\mathbf{D}.
\end{equation}
where $\mathbf{M}_{\Delta}$, $\mathbf{M}_{R}$ and $\mathbf{M}_\mathbf{D}$ represent depolarization, retardance, and diattenuation, respectively.

Diattenution $D$ quantifies the magnitude of reflection/transmission of polarized light and is defined as:
\begin{equation}
    D = \frac{1}{\mathbf{M}_{00}}\sqrt{\mathbf{M}_{01}^2 + \mathbf{M}_{02}^2 + \mathbf{M}_{03}^2}
\end{equation}

Polarizance $P$ quantifies the magnitude of unpolarized light that becomes polarized after reflection/transmission as follows:
\begin{equation}
    P = \frac{1}{\mathbf{M}_{00}}\sqrt{\mathbf{M}_{10}^2 + \mathbf{M}_{20}^2 + \mathbf{M}_{30}^2}
\end{equation}

Polarization preservation $\rho$ shows the preservation magnitude of the polarized incident  as follows:
\begin{equation}
    \rho = \frac{1}{3}(\mathbf{M}_{\Delta, 11} + \mathbf{M}_{\Delta, 20} + \mathbf{M}_{\Delta, 11})
\end{equation}
where higher $\rho$ values are typically associated with specular reflection components.

Retardance $R$ represents the phase shift introduced by retarder components and is given by:
\begin{equation}
    R = \cos^{-1} \left(\frac{\mathrm{tr}(\mathbf{M}_{R})}{2} - 1 \right)
\end{equation}
where \text{tr($\cdot$)} denotes the trace operator. To enhance the visualization of this phase information, we modulate the retardance with polarization preservation values, highlighting regions with notable polarization effects.

\newcommand{\captiondetailed}[3]{Detailed visualization of Mueller matrix of the #1 scene. The ROI is #2, and the average event rate is #3~MEv/s.}

\begin{figure}[h]
    \centering
    \includegraphics[width=\linewidth]{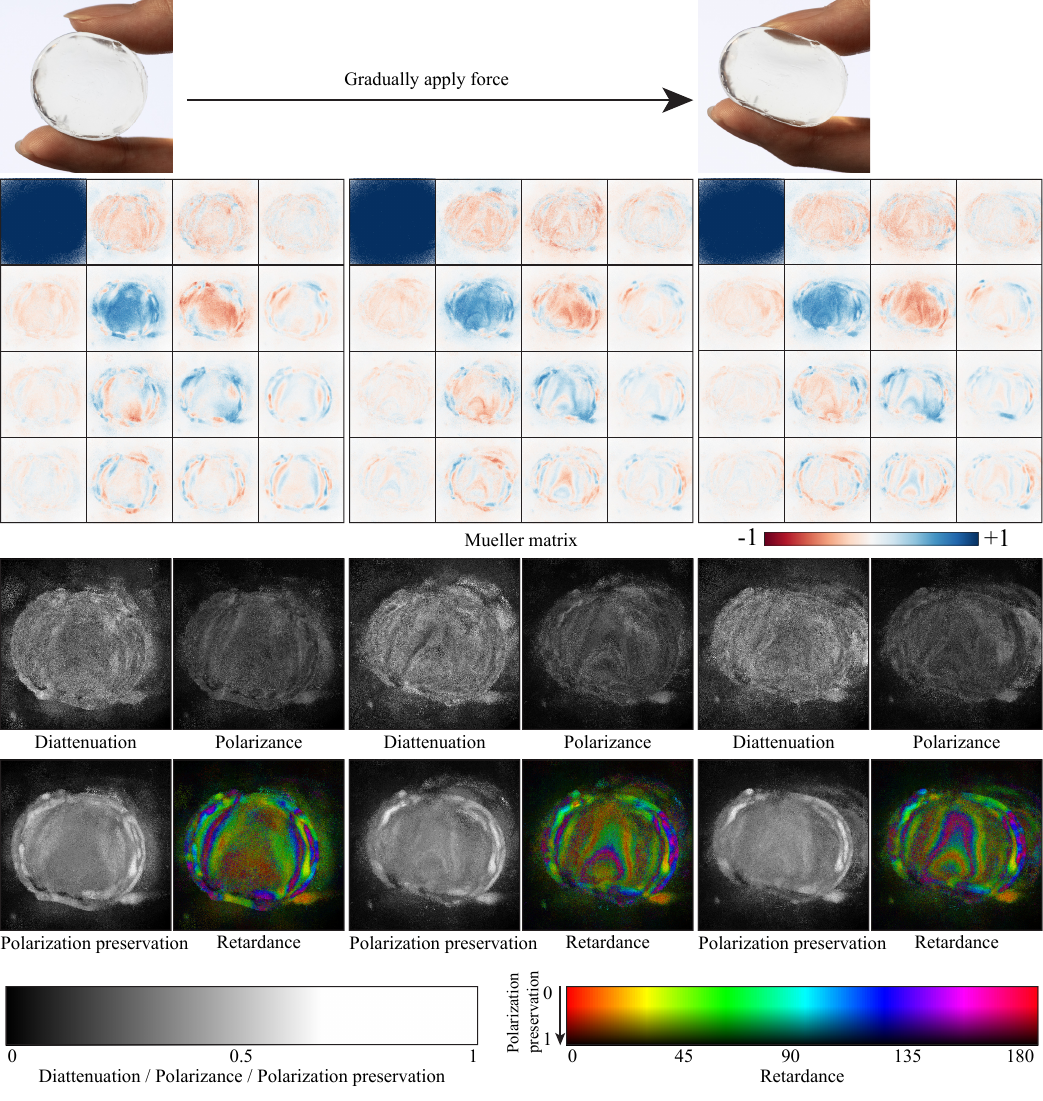}
    \caption{\captiondetailed{photoelasticity analysis}{500$\times$500}{151.7}}
    \label{fig:result_photoelasticity_supp}
\end{figure}

\newpage

\begin{figure}[h]
    \centering
    \includegraphics[width=\linewidth]{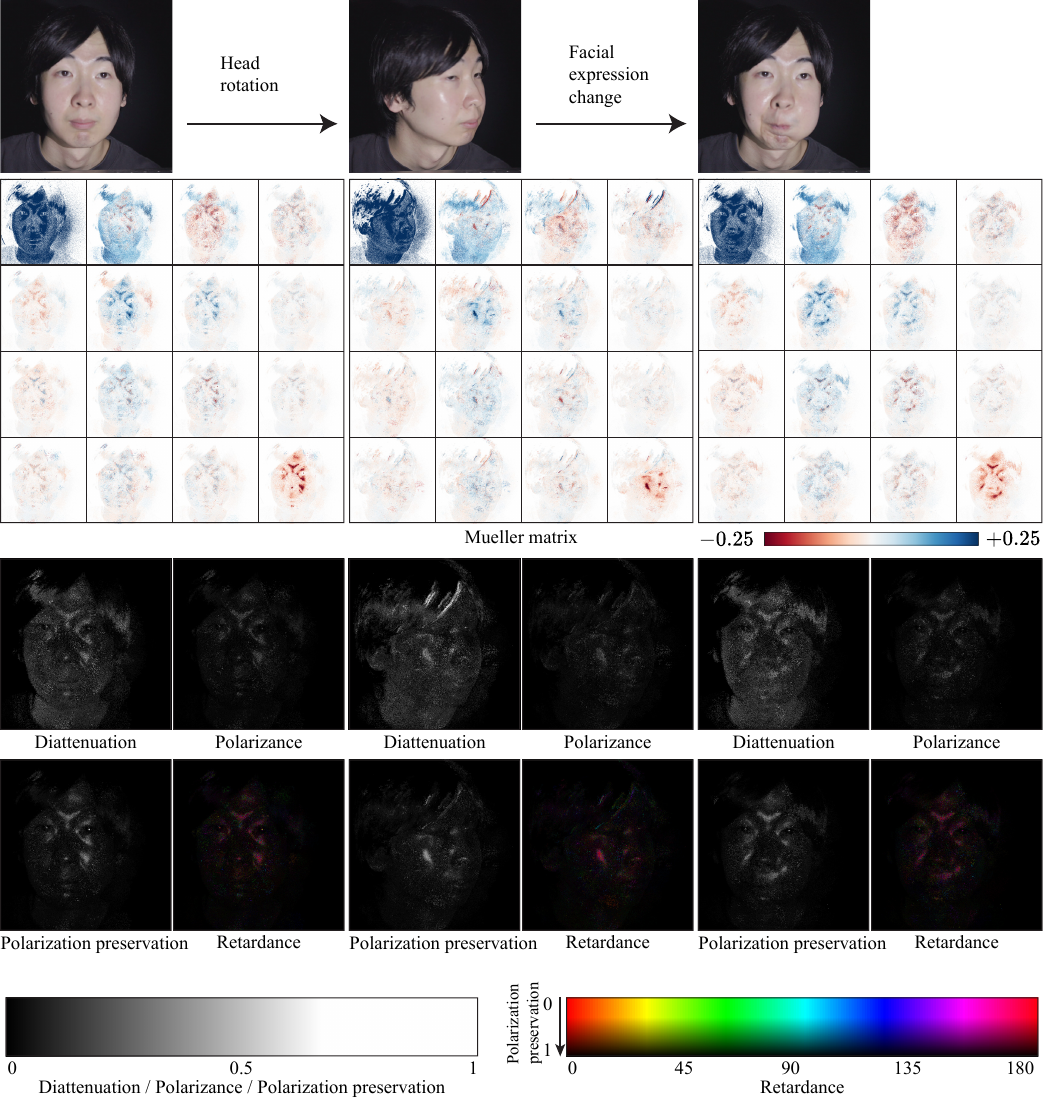}
    \caption{\captiondetailed{face}{600$\times$600}{26.6}}
    \label{fig:result_human_face}
\end{figure}

\newpage

\begin{figure}[h]
    \centering
    \includegraphics[width=\linewidth]{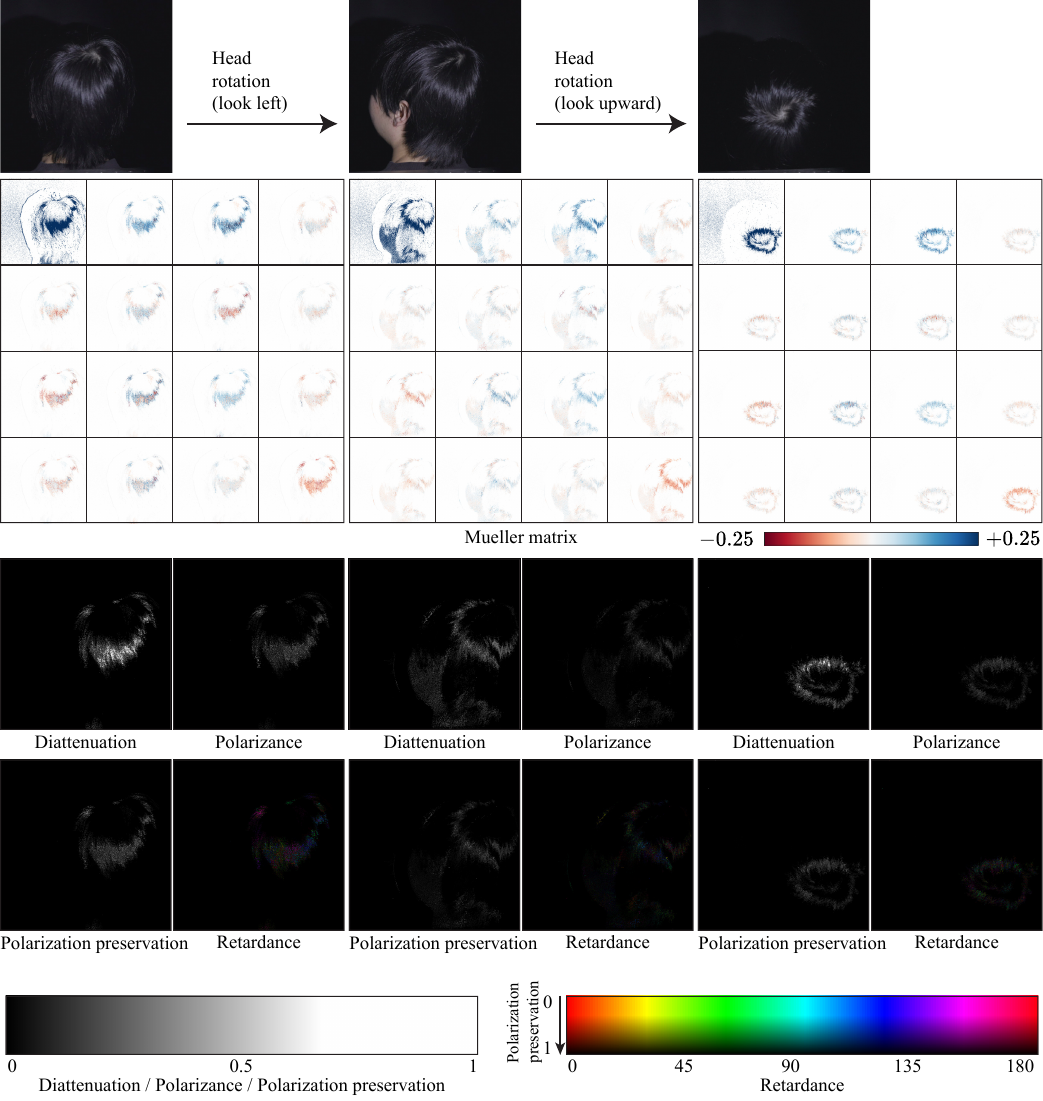}
    \caption{\captiondetailed{hair}{600$\times$600}{10.3}}
    \label{fig:result_human_hair}
\end{figure}

\newpage

\begin{figure}[h]
    \centering
    \includegraphics[width=\linewidth]{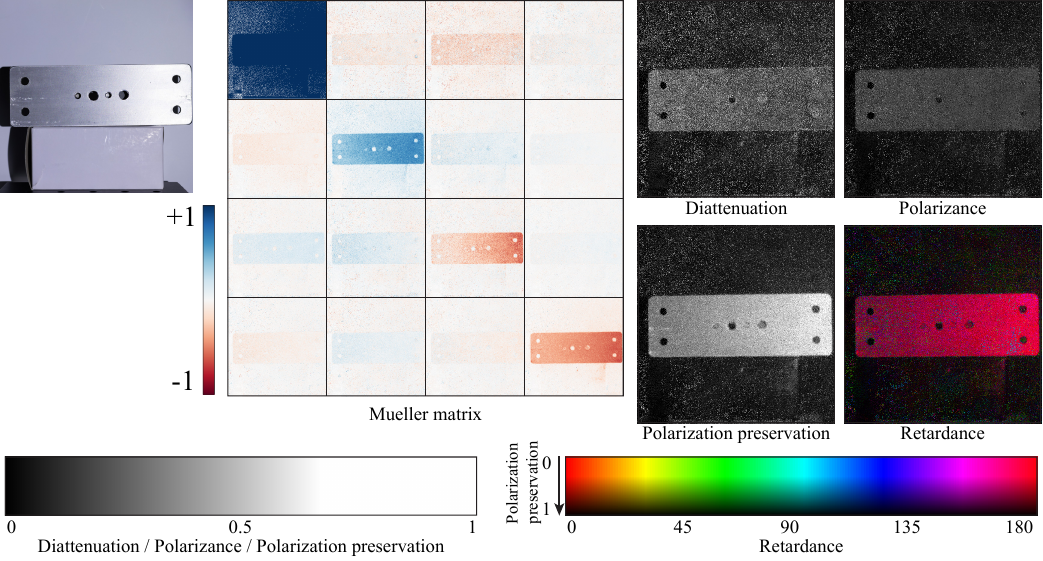}
    \caption{\captiondetailed{metal plate}{400$\times$400}{110.4}}
    \label{fig:result_metal_supp}
\end{figure}

\begin{figure}[h]
    \centering
    \includegraphics[width=\linewidth]{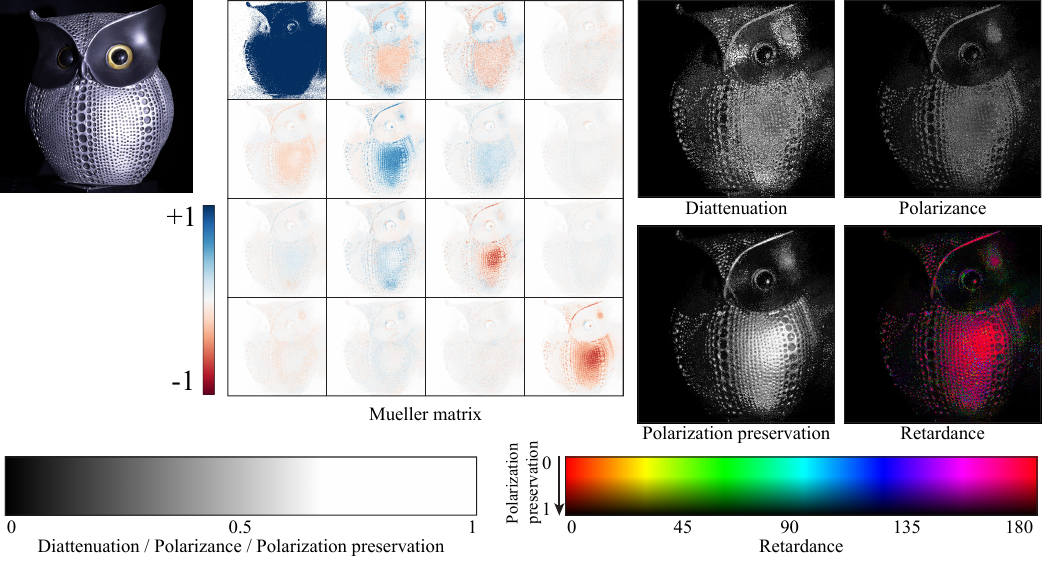}
    \caption{\captiondetailed{owl statue}{400$\times$400}{62.7}}
    \label{fig:result_owl_supp}
\end{figure}

\newpage

\begin{figure}[h]
    \centering
    \includegraphics[width=\linewidth]{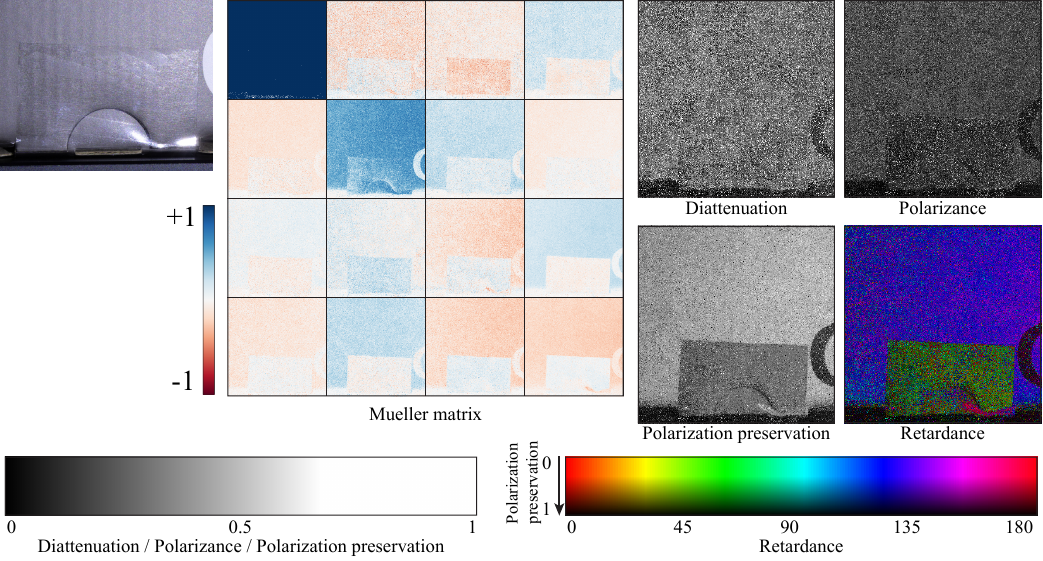}
    \caption{\captiondetailed{tape}{256$\times$256}{74.4}}
    \label{fig:result_tape_supp}
\end{figure}

\section{Additional Discussion}

\subsection{Wavelength Dependency}
Our prototype utilizes a white LED light source and a monochrome event camera, which captures a mixture of visible wavelengths. However, polarization states are wavelength-dependent, introducing potential inaccuracies. For instance, the QWP exhibits slight wavelength-dependent variations in retardance, and the reflection/transmission properties of the scene can also differ across wavelengths~\cite{jeon2024spectral}. To address this issue, a bandpass filter with the desired wavelength can be attached in front of the camera. This filter restricts light to a single wavelength, mitigating the wavelength dependency. However, this approach reduces the overall light intensity, potentially leading to a non-ideal response from the event sensor~\cite{gallego2020event, liu2024seeing}.

\subsection{Temporal Artifact in the Mueller-matrix Video}
The reconstructed Mueller matrix video exhibits frame-to-frame temporal artifacts. These artifacts are primarily caused by reflections on the QWP films due to the lack of adequate anti-reflection coatings and the presence of fine dust on the surface. Furthermore, the thin QWP films tend to fluctuate during rotation, causing variations in their surface normals and introducing temporal artifacts. 

\subsection{Motion Artifacts}
Our method assumes that small motions within a single frame are negligible. However, reconstruction accuracy may degrade when event generation is dominated by object motion rather than polarization-induced changes.
We observed such motion artifacts in the human hair scene (Figure~\ref{fig:result_human_hair}). Specifically, the dark regions of human hair rarely produce polarization-triggered events, making it challenging to accurately reconstruct the Mueller matrix. Addressing these limitations remains a topic for future work.

\subsection{Motor Selection}
In our hardware prototype, we utilized brushed DC motors, though alternative motor types could be considered. The brushed DC motor is affordable and offers high rotation speed under light load conditions. However, synchronizing two brushed DC motors precisely at high rotation speeds is difficult. We addressed this problem by independently controlling the rotation speed (Section~\ref{sec:rotation_speed_control}) and compensating for the phase difference between two motors using encoder signals (Section~\ref{sec:angle_time_correspondence}). 
For faster rotation speed and more precise control, a brushless DC motor, also known as an electronically commutated (EC) motor, presents a viable alternative. EC motors provide precise control at high speed, potentially simplifying the hardware system and improving rotation stability. However, EC motors are generally more expensive than brushed motors.

{\small
\bibliographystyle{ieeenat_fullname}
\bibliography{references}
}

\end{CJK}